\pdfoutput=1

\documentclass[11pt]{article}

\usepackage[final]{coling}

\usepackage{times}
\usepackage{latexsym}
\usepackage{url}
\usepackage{booktabs}
\usepackage{graphicx}
\usepackage{tabularx}
\usepackage{multirow}
\usepackage{pifont}
\usepackage{CJKutf8}
\newcommand{\textjp}[1]{\begin{CJK}{UTF8}{min}#1\end{CJK}} %

\usepackage[T1]{fontenc}

\usepackage[utf8]{inputenc}

\usepackage{microtype}

\usepackage{inconsolata}

\usepackage{graphicx}

\usepackage{CJKutf8}

\title{JMedBench: A Benchmark for Evaluating Japanese\\Biomedical Large Language Models}

\author{
    Junfeng Jiang\textsuperscript{\dag}\quad
    Jiahao Huang\textsuperscript{\dag}\quad
    Akiko Aizawa\textsuperscript{$\mathsection$\dag} \\
    \textsuperscript{\dag}The University of Tokyo\quad
    \textsuperscript{$\mathsection$}National Institute of Informatics\\
    \texttt{jiangjf@is.s.u-tokyo.ac.jp}\\
    \texttt{jiahao-huang@g.ecc.u-tokyo.ac.jp}\\
    \texttt{aizawa@nii.ac.jp}
}

\begin{document}
\maketitle
\begin{abstract}
Recent developments in Japanese large language models (LLMs) primarily focus on general domains, with fewer advancements in Japanese biomedical LLMs. One obstacle is the absence of a comprehensive, large-scale benchmark for comparison. Furthermore, the resources for evaluating Japanese biomedical LLMs are insufficient. To advance this field, we propose a new benchmark including eight LLMs across four categories and 20 Japanese biomedical datasets across five tasks. Experimental results indicate that: (1) LLMs with a better understanding of Japanese and richer biomedical knowledge achieve better performance in Japanese biomedical tasks, (2) LLMs that are not mainly designed for Japanese biomedical domains can still perform unexpectedly well, and (3) there is still much room for improving the existing LLMs in certain Japanese biomedical tasks. Moreover, we offer insights that could further enhance development in this field. Our evaluation tools tailored to our benchmark as well as the datasets are publicly available to facilitate future research.\footnote{\url{https://huggingface.co/datasets/Coldog2333/JMedBench}} 
\end{abstract}

\section{Introduction}
Large language models (LLMs) show excellent performances in various tasks in general domains including Question Answering (QA) \cite{brown2020language,alpaca}, Machine Translation (MT) \cite{he2024exploring}, Summarization \cite{ravaut2024context}, Machine Reading Comprehension (MRC) \cite{zhou2023context}, 
Sentiment Analysis \cite{zhang2024sentiment}, and so on. Some researchers design proper prompts for solving biomedical tasks \cite{singhal2023large,lievin2024can,nori2023can}. However, most of the existing LLMs have been pre-trained with texts in general domains, lacking domain-specific knowledge. To overcome this challenge, biomedical LLMs are proposed through pre-training on biomedical corpora \cite{chen2023meditron,wu2024pmc}, fine-tuning with instruction data \cite{han2023medalpaca}, or reinforcement learning with human feedback \cite{yang2024zhongjing}. 

With the chain-of-thought prompting technique, \citet{lievin2024can} have achieved 60.2\% accuracy on USMLE-QA \cite{jin2021disease}, passing the medical licensing examination in the United States. In the most recent work, with the help of multiple agents, \citet{nori2023can} have achieved 93.06\% accuracy on the USMLE-QA dataset, similar to the performance of a human expert. With this series of techniques, biomedical LLMs are greatly promoted in English biomedical tasks. However, biomedical LLMs in other languages still have much room for improvement (e.g., Japanese, Chinese, French, etc.). Besides the relative unpopularity of existing Japanese LLMs, another important obstacle is the lack of a comprehensive benchmark for evaluation and comparison. Therefore, in this paper, we focus on constructing a benchmark for evaluating Japanese biomedical LLMs.

We selected five tasks that are widely used for evaluating LLMs and real-world applications, including multi-choice question-answering (MCQA), named entity recognition (NER), machine translation (MT), document classification (DC), and semantic text similarity (STS). Since there are only a few Japanese biomedical datasets exist and they are generally small (e.g., IgakuQA \cite{kasai2023evaluating} only has 1,600 samples for testing), to reduce the fluctuation of evaluation results, we translate large-scale and high-quality datasets from other languages (e.g., English) to Japanese, augmenting the scale of our benchmark. Furthermore, in the field of Japanese biomedical LLM, a solid leaderboard is missing. Therefore, we select eight representative models to conduct extensive experiments, providing a standard for comparison. We hope our work can make future comparisons more convenient and fair, promoting the development in this field.

\begin{table*}[tbh]
  \tabcolsep=3.5pt
  \centering
  \small
  \renewcommand{\arraystretch}{1.05}
      \begin{tabular}{|p{4.5cm}|cc|cc|cc|c|}
        \toprule
            \multirow{2}{*}{\textbf{Benchmark}} & \multirow{2}{*}{\textbf{Language}} & \multirow{2}{*}{\textbf{Domain}} & \multicolumn{2}{c|}{\textbf{Task}} & \multirow{2}{*}{\textbf{\#Dataset}} & \multirow{2}{*}{\textbf{\#Sample}} & \multirow{2}{*}{\textbf{Creator}} \\
            &  &  & MCQA & Others &  & & \\
        \midrule
          BLURB \cite{gu2021domain}  & English & Biomedical & \ding{51} & \ding{51} & 13 & 65,146 & Human  \\  
          MMLU \cite{chang2024survey}  & English & Mixed  & \ding{51} &  \ding{55} & 1 & 14,042 & Human  \\  
          JMMLU  & Japanese & Mixed  & \ding{51}  & \ding{55} & 1 & 7,097 & Translation \\  
          DrBenchmark \cite{labrak-etal-2024-drbenchmark} & French & Biomedical  & \ding{51} & \ding{51} & 20 & 10,519 & Human \\  
          MMedBench \cite{qiu2024towards} & Multilingual & Biomedical  & \ding{51} & \ding{55} & 6 & 8,518 & Human \\  
          JMedBench & Japanese & Biomedical  & \ding{51} & \ding{51} & 20 & 38,130 &  Mixture \\ 
        \bottomrule
      \end{tabular}
  \caption{Comparison of existing benchmarks.}
  \label{tab:comparison_benchmark}
\end{table*}

In summary, our contributions are in three folds.
\begin{itemize}
    \item We construct a large-scale benchmark including 20 Japanese biomedical datasets across five tasks for a comprehensive evaluation.
    \item We evaluate eight representative models across four categories in our benchmark to provide a standard for future comparison.
    \item We conduct extensive analysis from aspects of the dataset, model, and prompt template, providing valuable insights for future researchers.
\end{itemize}

\section{Related Works}
Benchmarking is essential for the development of a specific field. ImageNet Challenge \cite{deng2009imagenet} is a famous benchmark in Computer Vision. Many remarkable works on image recognition have been proposed \cite{krizhevsky2012imagenet,he2016deep,tan2019efficientnet} throughout history and the development has increased rapidly. One reason for this success is the convenience of comparison and evaluation in this field. The GLUE \cite{wang2018glue} is another famous benchmark for evaluating and analyzing natural language understanding (NLU) systems to promote research in developing general and robust NLU systems. However, these works mainly focus on English tasks, limiting the scope of evaluating other languages like Japanese. \citet{kurihara2022jglue} constructed the JGLUE from scratch without using any translation, including six datasets,
which facilitates the research in Japanese natural language processing (NLP) \cite{yano-etal-2024-multilingual,enomoto2024investigating,aizawa2024llm}.

Considering the wide applications of language models (LMs), researchers are trying to explore LMs' power in biomedical tasks. \citet{gu2021domain} collected 13 biomedical NLP datasets in six tasks from different isolated work to form a benchmark called BLURB for evaluating biomedical models. MMLU \cite{chang2024survey} is a benchmark consisting of multiple topics. Specially, it contains some biomedical questions like medical questions at the college level. DrBenchmark \cite{labrak-etal-2024-drbenchmark} is an NLU benchmark for evaluating French biomedical models. However, they are not applicable in Japanese. JMMLU\footnote{\url{https://github.com/nlp-waseda/JMMLU}} is a translated version of the MMLU. The researchers recruited human translators to check and remove those that were difficult to translate, irrelevant, or inconsistent with the Japanese culture. Recently, \citet{qiu2024towards} have proposed a multilingual benchmark with six languages for evaluating medical LMs. These benchmarks reflect some shortages of existing LLMs and provide insights into improving the Japanese biomedical LLMs, but they only focus on the MCQA tasks, which hinders the completeness of the evaluation. Considering these shortages, in this paper, we are dedicated in constructing a large-scale benchmark with diverse tasks for evaluating Japanese biomedical large language models. Table \ref{tab:comparison_benchmark} shows a comparison of these benchmarks.





\begin{figure}[thb]
    \centering
    \includegraphics[width=\linewidth]{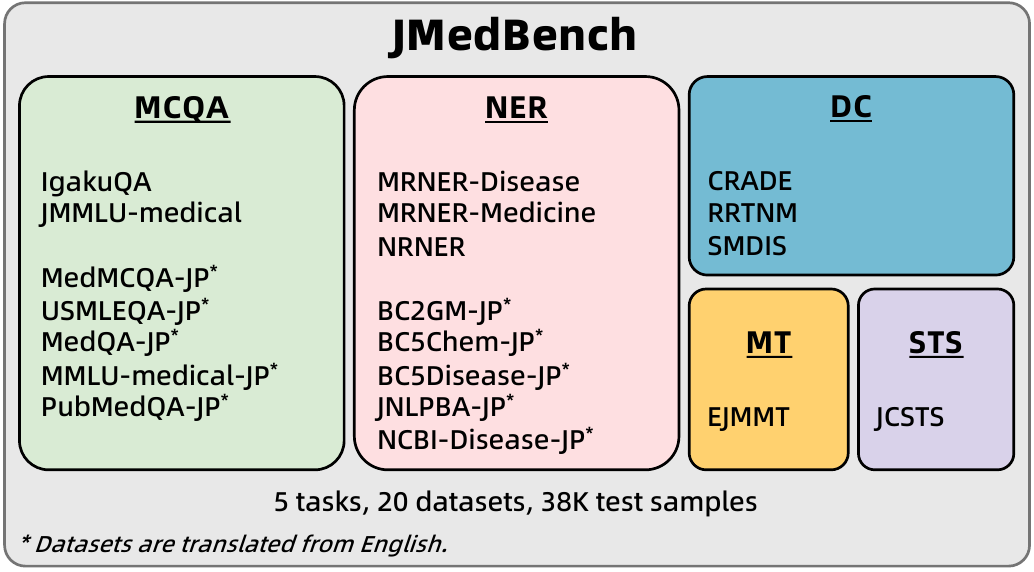}
    \caption{Overview of JMedBench}
    \label{fig:overview}
\end{figure}

\section{JMedBench}
Our benchmark construction consists of two parts. The first part is the dataset collection, while another part is the protocol for evaluation. Firstly, we introduce the rationality of dataset selection and how we augment our benchmark with datasets from other languages. Then, we propose a protocol to obtain robust evaluation results and discuss its necessity for evaluating Japanese biomedical LLMs. Figure \ref{fig:overview} is an overview of our benchmark.

\subsection{Datasets}
In the JMedBench, we include 20 datasets across five tasks containing 38K testing samples. We collect some human-manufactured Japanese datasets, like IgakuQA \cite{kasai2023evaluating}. We also translate some high-quality large-scale English datasets into Japanese to enhance the robustness of JMedBench. Considering the convenience and performance of using OpenAI's API, we use ChatGPT\footnote{\url{https://openai.com/index/chatgpt/}} and GPT-4 \cite{achiam2023gpt} to create our evaluation datasets when translation is needed. To ensure the quality of the translated testing sets, we use the most powerful model from OpenAI, the GPT-4\footnote{We used \texttt{gpt-4-0613} checkpoint.}, to perform machine translation. In-context learning is a common practice for adapting an LLM to an unseen task. Therefore, we also translate the training or validation sets to support few-shot evaluation. Due to the limitation of our budgets, we use the cheapest API\footnote{We used \texttt{gpt-3.5-turbo-1106} checkpoint.} from OpenAI to translate these samples. Though the translation may not be perfect, producing unfaithful content sometimes, it is good enough to provide information like some domain-specific knowledge and task format during the few-shot evaluation. Previous works \cite{hendy2023good,sanz2023google,alafnan2024large} also have similar findings that ChatGPT has already had a comparable MT performance with specialized Neural Machine Translation systems. Here listed are the involved biomedical tasks and corresponding datasets. Detailed statistics can be found in Table \ref{tab:dataset_statistics} in the Appendix.

\begin{itemize}
    \item 
    \textbf{MCQA} is one of the most commonly used tasks for evaluating LLMs since other tasks can be easily reformulated into the MCQA task. We included IgakuQA \cite{kasai2023evaluating}, JMMLU-medical\footnote{\url{https://github.com/nlp-waseda/JMMLU}}, and translated MedMCQA \cite{pmlr-v174-pal22a}, MedQA \cite{jin2021disease}, USMLE-QA, PubMedQA \cite{jin-etal-2019-pubmedqa}, and MMLU-medical \cite{hendryckstest2021,hendrycks2021ethics}.

    \item 
    \textbf{MT} is an important natural language generation (NLG) task. In the biomedical domain, researchers usually need to refer to some English terminologies or communicate with other researchers. Therefore, we expect LLMs can handle cross-lingual tasks besides monolingual tasks. We included the EJMMT \cite{hayakawa2020fine} dataset to evaluate the cross-lingual ability of LLMs.
    
    \item 
    \textbf{NER} is an NLU task aiming to extract named entities like biomedical terminologies, medicines, etc. We included three Japanese medical NER datasets from JMED-LLM\footnote{\url{https://github.com/sociocom/JMED-LLM}}: MRNER-disease, MRNER-medicine, and NRNER. To improve the diversity of the dataset, we also follow the BLURB benchmark and include translated BC2GM \cite{smith2008overview}, BC5Chem, BC5-Disease \cite{li2016biocreative}, JNLPBA \cite{collier2004introduction}, and NCBI Disease \cite{dougan2014ncbi}. 
    

    \item 
    \textbf{DC} aims to classify documents into specific categories. We include three datasets from JMED-LLM: CRADE, RRTNM, and SMDIS. 

    \item 
    \textbf{STS} is a regression task to compute the semantic similarity between two biomedical sentences. We reformulate it as a classification task to output the discrete level of similarity. We include the JCSTS \cite{mutinda2021semantic}. 
    
\end{itemize}

\subsection{Evaluation Dataset Augmentation}
To enlarge the size of JMedBench for obtaining robust evaluation results, we select several biomedical datasets in English, because of its popularity.

\subsubsection{Multi-choice Question-Answering}
\label{sec:mcqa_data_augmentation}
Different from previous works 
that usually conduct machine translation at the sentence level, we perform translation at the instance level. Specifically, we translate questions and options meanwhile, so that LLM can understand the scenario better to provide more correct translations. Detailed prompt template can be found in Table \ref{tab:augment_prompt_mcqa} in the Appendix.

\subsubsection{Named Entity Recognition}
We also translate the NER datasets from the BLURB benchmark to improve the amount and diversity of JMedBench. There are three fields in the NER samples: entity type, text, and entities. To ensure the consistency of the translated entity types, we manually translate them into Japanese based on a dictionary (e.g., gene $\rightarrow$\textjp{遺伝子}). As for the text and entities, we also perform translation at the instance level, as described in Section \ref{sec:mcqa_data_augmentation}. The prompt template for translating the biomedical NER datasets is also shown in Table \ref{tab:augment_prompt_ner} in the Appendix. 

One of the challenges is that the translated entities may not appear in the translated text. To solve this issue, we conduct the translation in two phases: machine translation and manual modification. We first use ChatGPT and GPT-4 to translate the training and testing sets, respectively. We then collect all the invalid samples, mainly due to JSON format error and failure to include the translated entities, and re-translate them using GPT-4. We increase the temperature to 0.5 and call the GPT-4 API again at most 5 times to seek a valid sample. After the machine translation phase, 223 translated entries (0.34\%) still remain invalid and then we manually modify these entries to make them valid. 

During machine translation, we find that translating entities first instead of text first can reduce about 10\% of invalid samples. We speculate that with the entity-first prompt, LLM can refer to the already translated entities when translating the text, thus, the translated entities are usually contained in the following translated sentence.
However, since this is not the main focus of this paper, we did not conduct further analysis to verify this hypothesis. We hope this finding can inspire future researchers when performing instance-level machine translation. Despite a small number of failure cases during the machine translation phase (some bad cases can be found in the Appendix \ref{app:badcase_ner}), we realized that the translation quality is very high when we conduct manual modification, which also reflects the reliability of our data augmentation method.

\begin{table*}[!tbh]
  \tabcolsep=3.5pt
  \centering
  \small
  \renewcommand{\arraystretch}{1.05}
      \begin{tabularx}{\textwidth}{|p{2.5cm}|>{\centering\arraybackslash}X>{\centering\arraybackslash}X>{\centering\arraybackslash}X>{\centering\arraybackslash}X>{\centering\arraybackslash}X>{\centering\arraybackslash}X>{\centering\arraybackslash}X|>{\centering\arraybackslash}p{2cm}|}
        \toprule
          \textbf{Accuracy (\%)}  & \textbf{IGA} 
 & \textbf{JMM}  & \textbf{MedM} & \textbf{USM}  & \textbf{MedQ}  &  \textbf{MML} &  \textbf{Pub} & \textbf{Aver (Micro)} \\
        \midrule
          \multicolumn{9}{|p{5cm}|}{\textbf{Zero-shot Evaluation}} \\
        \midrule
          Llama2-7B & 22.69 & 26.20 & 30.31 & 27.81 & 23.17 & 29.77 & 63.50 & 30.91 \\
          Llama3-8B & 26.19 & 35.09 & 31.94 & 32.21 & 26.00 & 36.77 & 62.30 & 34.51 \\  
          Qwen2-7B &  \textbf{41.25} & \textbf{44.06} & \textbf{38.03} & \textbf{38.49} & \textbf{31.03} & \textbf{49.01} & 68.90 & \textbf{42.58}  \\  
          Mistral-7B &  25.19 & 30.68 & 30.60 & 28.44 & 23.57 & 32.82 & 68.80 & 32.74 \\
          Meditron-7B & 21.94 & 25.65 & 28.31 & 26.39 & 21.92 & 25.65 & 56.50 & 28.56 \\  
          llm-jp-13B & 31.00 & 36.51 & 30.46 & 31.66 & 25.29 & 35.54 & \textbf{73.60} & 35.17 \\ 
          SwallowLM-7B &  27.88 & 29.50 & 29.26 & 27.73 & 22.39 & 29.88 & \underline{70.70} & 31.86 \\  
          MMed-Llama3-8B & \underline{35.56} & \underline{37.45} & \underline{35.43} & \underline{36.92} & \underline{29.54} & \underline{38.86} & 70.00 & \underline{38.64} \\  
        \midrule     
          \multicolumn{9}{|p{5cm}|}{\textbf{Few-shot Evaluation}} \\
        \midrule
          Llama2-7B & 23.56 & 29.35 & 29.95 & 29.07 & 24.43 & 32.71 & 55.80 & 31.28 \\  
          Llama3-8B & 36.31 & 37.77 & 36.77 & 35.04 & 29.30 & 43.77 & \underline{72.50} & 39.97 \\  
          Qwen2-7B & \textbf{51.75} & \textbf{51.61} & \textbf{42.74} & \textbf{42.42} & \textbf{35.51} & \textbf{61.04} & \underline{72.50} & \textbf{49.03} \\  
          Mistral-7B & 30.31 & 33.60 & 31.80 & 29.62 & 23.96 & 37.20 & 72.40 & 35.07 \\
          Meditron-7B &  22.31 & 28.25 & 28.57 & 27.73 & 24.19 & 28.92 & 55.80 & 29.80 \\  
          llm-jp-13B & 36.06 & 37.37 & 32.54 & 33.62 & 26.32 & 39.44 & \textbf{75.90} & 37.54 \\
          SwallowLM-7B & 29.00 & 33.67 & 32.23 & 30.32 & 23.41 & 37.89 & 71.40 & 35.16 \\  
          MMed-Llama3-8B & \underline{45.37} & \underline{46.42} & \underline{38.54} & \underline{41.95} & \underline{34.88} & \underline{50.29} & \underline{72.50} & \underline{44.64} \\  
        \bottomrule
      \end{tabularx}
  \caption{Benchmark results on Japanese biomedical MCQA tasks, including IgakuQA (\textbf{IGA}) and JMMLU-medical (\textbf{JMM}), as well as the translated versions of MedMCQA (\textbf{MedM}), USMLE-QA (\textbf{USM}), MedQA (\textbf{MedQ}), MMLU-medical (\textbf{MML}), and PubMedQA (\textbf{Pub}). We report the highest accuracy among four prompt templates as discussed in Section \ref{sec:experimental_settings}. The best and second-best performances are highlighted in bold and underlined, respectively.}
  \label{tab:benchmark_mcqa}
\end{table*}

\subsection{Evaluation Protocols}
\label{sec:experimental_settings}
LLMs are usually sensitive to the prompt templates, especially in zero-shot evaluation \cite{gan-mori-2023-sensitivity}. 
To obtain a robust and fair result, we suggest reporting the maximal score of multiple runs using diverse prompt templates for benchmarking. We have also considered computing an average score using different templates, whereas this reported performance may be easily implicated by inappropriate prompts (e.g., using an English-centric prompt for a Japanese-only LLM). In the following evaluation, we use four types of prompt templates:
\begin{itemize}
    \item \textbf{Minimal}: We include information as little as possible in the prompt. For example, for completing the MCQA task, we only input the question, and compute the likelihood of each possible option, namely, \texttt{\{question\}\textbackslash n}.
    
    \item \textbf{Standard}: We use commonly used prompt templates in each task. For example, we follow \cite{robinson2023leveraging} for evaluating MCQA tasks.

    \item \textbf{English-centric}: Some of the existing Japanese LLMs were continually pre-trained from English-centric LLMs. Therefore, we intend to explore whether an English-centric prompt template is beneficial.
    
    \item \textbf{Instructed}: Besides the standard input, we include a brief task instruction, evaluating the instruction-following ability of LLMs.
\end{itemize}

As for the MCQA and DC tasks, it is difficult to constrain the auto-regressive LLMs to generate one of the given options or classes. Therefore, we follow \citet{eval-harness} to compute the likelihood perplexity of each possible answer and select the one that has the highest generation possibility as the final answer. We report accuracy on these two tasks. As for the STS task, we also calculate the likelihood perplexity of generating 0-5 and select the one that has the highest generation possibility as the final output. We use the Pearson Correlation as the evaluation metric.
As for the MT and NER tasks, we generate the output and compute the BLEU \cite{papineni2002bleu} score and entity-level F1 score, respectively.

\section{Experiments}

\subsection{Comparison Methods}
In our experiments, we included four categories of popular and excellent LLMs to construct our benchmark, including 
\textbf{general LLMs in other languages}: Llama2 \cite{touvron2023llama}, Llama3 \cite{dubey2024llama}, Qwen-2 \cite{yang2024qwen2}, Mistral \cite{jiang2023mistral}; 
\textbf{biomedical LLM in other languages}: Meditron \cite{chen2023meditron}; 
\textbf{Japanese general LLMs}: llm-jp \cite{aizawa2024llm}, SwallowLM \cite{fujii2024continual}; 
and \textbf{Japanese biomedical LLM}: MMed-Llama3 \cite{qiu2024towards}. 
The specific checkpoints are listed in Table \ref{tab:model_info} in the Appendix. Due to the computation resources, we only evaluate LLMs with around $7\sim8$B parameters. Llm-jp is a representative LLM that was pre-trained from scratch with Japanese and English texts. Although it does not have the 7B version of the model, we still include the llm-jp with 13B parameters in our benchmark.





    

\begin{table*}[!tbh]
  \tabcolsep=3.5pt
  \centering
  \small
  \renewcommand{\arraystretch}{1.05}
      \begin{tabularx}{\textwidth}{|p{2.5cm}|>{\centering\arraybackslash}X>{\centering\arraybackslash}X>{\centering\arraybackslash}X>{\centering\arraybackslash}X>{\centering\arraybackslash}X>{\centering\arraybackslash}X>{\centering\arraybackslash}X>{\centering\arraybackslash}X|>{\centering\arraybackslash}p{2cm}|}
        \toprule
          \textbf{F1-entity (\%)}  & \textbf{MRD} 
 & \textbf{MRM}  & \textbf{NRN} & \textbf{B2G}  & \textbf{B5C}  &  \textbf{B5D} &  \textbf{JNL} & \textbf{NCB} & \textbf{Aver (Micro)} \\
        \midrule
          \multicolumn{10}{|p{5cm}|}{\textbf{Zero-shot Evaluation}} \\
        \midrule
          Llama2-7B & 0.74 & 18.99 & 10.12 & 32.37 & 58.74 & 38.33 & 7.76 & 36.21 & 34.74  \\  
          Llama3-8B &3.57 & 18.43 & \underline{14.97} & 36.17 & 58.67 & \underline{40.91} & \textbf{24.69} & \textbf{52.70} & \textbf{40.69}  \\   
          Qwen2-7B & 3.06 & 15.02 & 9.54 & \textbf{39.88} & 52.26 & 38.40 & 8.51 & 40.13 & 35.43  \\  
          Mistral-7B & \textbf{16.75} & \textbf{30.21} & 11.33 & 35.61 & 52.37 & 38.92 & 7.12 & 46.65 & 34.65  \\  
          Meditron-7B & 1.94 & 4.78 & 5.17 & 15.31 & 31.12 & 17.71 & 12.89 & 18.29 & 19.14  \\  
          llm-jp-13B & \underline{8.80} & 11.99 & 14.58 & 29.31 & \underline{59.15} & 37.62 & 22.52 & 43.55 & 37.41  \\  
          SwallowLM-7B & 2.20 & 23.74 & 11.79 & 31.18 & 58.22 & \textbf{41.76} & 13.22 & 34.85 & 36.26  \\  
          MMed-Llama3-8B & 3.77 & \underline{26.85} & \textbf{17.25} & \underline{39.70} & \textbf{61.85} & 39.21 & \underline{16.48} & \underline{51.33} & \underline{40.18}  \\  
        \midrule     
          \multicolumn{10}{|p{5cm}|}{\textbf{Few-shot Evaluation}} \\
        \midrule
          Llama2-7B &11.10 & 21.14 & 20.41 & 46.76 & 72.95 & 55.50 & 47.85 & 52.90 & 55.22 \\  
          Llama3-8B & \underline{15.83} & \underline{37.26} & 25.15 & \textbf{51.98} & \underline{79.42} & \underline{63.40} & \textbf{53.47}& \textbf{62.05} & \textbf{61.69} \\  
          Qwen2-7B &  11.65 & 22.31 & 24.93 & \underline{50.59} & 76.96 & 55.23 & 49.54 & 57.55 & 57.69  \\  
          Mistral-7B & 15.39 & 32.50 & \underline{26.31} & 48.15 & 73.06 & 56.12 & 48.11 & 51.33 & 55.83 \\  
          Meditron-7B & 10.70 & 18.73 & 19.13 & 45.12 & 68.36 & 52.05 & 46.02 & 52.49 & 52.47 \\  
          llm-jp-13B & 14.74 & 22.23 & 24.64 & 45.25 & 76.60 & 59.79 & \underline{51.77} & 56.14 & 57.76 \\  
          SwallowLM-7B & 12.05 & 25.58 & 20.55 & 44.41 & 74.74 & 59.26 & 46.60 & 51.03 & 55.62  \\  
          MMed-Llama3-8B & \textbf{17.27} & \textbf{39.47} & \textbf{29.09} & 49.19 & \textbf{80.34} & \textbf{65.27} & 51.05 & \underline{61.21} & \underline{61.14} \\ 
        \bottomrule
      \end{tabularx}
  \caption{Benchmark results on Japanese biomedical NER tasks, including MRNER-Disease (\textbf{MRD}), MRNER-Medicine (\textbf{MRM}) and NRNER (\textbf{NRN}), as well as the translated versions of BC2GM (\textbf{B2G}), BC5Chem (\textbf{B5C}), BC5Disease (\textbf{B5D}), JNLPBA (\textbf{JNL}), and NCBI-Disease (\textbf{NCB}). We report the highest F1-entity score among four prompt templates as discussed in Section \ref{sec:experimental_settings}. The best and second-best performances are highlighted in bold and underlined, respectively.}
  \label{tab:benchmark_ner}
\end{table*}

\subsection{Experimental Results}
\subsubsection{Multi-choice Question-Answering}
\label{sec:experiment_mcqa}
Table \ref{tab:benchmark_mcqa} shows the benchmark results on Japanese biomedical MCQA tasks. Surprisingly, Qwen2 outperforms all models in MCQA, followed by MMed-Llama3. Note that Qwen2 was primarily pre-trained with Chinese and English texts. We hypothesize that one reason for its success is the considerable overlap in tokens between Chinese and Japanese. MMed-Llama3 was continually pre-trained on biomedical texts in multiple languages including Japanese, explaining its superior performance over Llama3. These observations highlight the importance of understanding the Japanese language and injecting domain knowledge. With few-shot demonstrations, all models have improved. We attribute this to the task format \cite{min-etal-2022-rethinking} and some domain-specific knowledge provided by the demonstrations. Comparing Llama2 and Llama3, we find that the performance gap under the zero-shot setting is larger than that under the few-shot setting. The additional improvement should be attributed to the improved in-context learning (ICL) ability of Llama3, highlighting the need to enhance the ICL ability of LLMs. 

Although there is a human-translated version of MMLU-medical, namely, the JMMLU-medical dataset, we still translate the original MMLU-medical dataset using GPT-4 to enrich our benchmark. According to the performances of these two datasets (i.e., JMM \& MML in Table \ref{tab:benchmark_mcqa}), the differences between performances on these two datasets do not exceed 5\% of accuracy. Furthermore, the ranking of the performances on the translated MMLU-medical dataset also reflects the ranking on the human-translated JMMLU-medical dataset. These observations confirm the quality and the applicability of our translated datasets.

Meditron was continually pre-trained with large-scale English biomedical texts from the Llama2 checkpoint. \citet{chen2023meditron} showed that Meditron has been successfully shifted to the biomedical domain, outperforming the vanilla Llama2 in various biomedical MCQA tasks. However, we realize that Meditron performs worse than Llama2 
in the JMedBench. Such multilingual ability degradation is probably due to the catastrophic forgetting issue during continual pre-training. How to improve an LLM safely without losing any other ability should be considered in future research. Besides, since the SwallowLM and MMed-Llama3 were continually pre-trained with additional Japanese texts from Llama2 and Llama3, respectively, they are improved by approximately $1\%\sim5\%$ average accuracy, indicating the importance of local-language adaptation. 




\subsubsection{Named Entity Recognition}
Table \ref{tab:benchmark_ner} shows the results on Japanese biomedical NER datasets. In the few-shot evaluation of BC2GM, BC5Chem, BC5Disease, JNLPBA, and NCBI-Disease datasets, we use three shots of examples. However, for MRNER-Disease, MRNER-Medicine and NRNER, we only use one shot of example because texts in these datasets are so long that multiple shots will exceed the input token limit of several models. 

According to the results, we find that Llama3-8B outperforms other LLMs in both zero-shot and few-shot evaluations, with average F1-entity score of 40.69\% and 61.69\% respectively. The Japanese biomedical LLM, MMed-Llama3, has the second-best performance in both settings. Few-shot examples can significantly improve the performance of models on the NER tasks, ranging from 19.36\% to 33.33\% F1-entity improvement. Similar to the observations on MCQA tasks, we believe these examples help LLMs better understand the entity types' definition and output format.
Besides, we find that LLMs perform generally worse on datasets including MRNER-Disease, MRNER-Medicine, and NRNER which are derived from JMED-LLM. Note that the average text lengths of datasets from these two sources are 69.82 and 247.81 Japanese characters, while the numbers of entities are 1.33 and 2.66, respectively. Considering the longer input text, larger number of entities and sparser entity distribution, we believe these are the main reasons why the datasets derived from JMED-LLM are more challenging.

\begin{table*}[!tbh]
  \tabcolsep=3.5pt
  \centering
  \small
  \renewcommand{\arraystretch}{1.05}
      \begin{tabularx}{\textwidth}{|p{2.5cm}|>{\centering\arraybackslash}X>{\centering\arraybackslash}X>{\centering\arraybackslash}X|>{\centering\arraybackslash}X>{\centering\arraybackslash}X>{\centering\arraybackslash}X>{\centering\arraybackslash}X|>{\centering\arraybackslash}X|}
        \toprule
          \multirow{2}{*}{\textbf{Metric}}  & \textbf{EJMMT (en->ja)} & \textbf{EJMMT (ja->en)} 
 & \textbf{Aver} & \textbf{CRADE}  & \textbf{RRTNM} & \textbf{SMDIS}  & \textbf{Aver (Micro)} & \textbf{JCSTS} \\
          \cmidrule(lr){2-9}
            & \multicolumn{3}{c|}{\textbf{BLUE}} & \multicolumn{4}{c|}{\textbf{Accuracy (\%)}} &  \textbf{Pearson} \\ 
        \midrule
          \multicolumn{9}{|p{5cm}|}{\textbf{Zero-shot Evaluation}} \\
        \midrule
          Llama2-7B & 11.13  & 14.18  & 12.65 & 27.17 & 37.08 & 54.76  & 39.67 & -0.005  \\  
          Llama3-8B & 16.79	 &  \textbf{23.66}    & \underline{20.23} & 25.00 & 44.94 & 51.19  & 40.38 & 0.422   \\ 
          Qwen2-7B & 15.24  &  19.59   & 17.41 & \textbf{35.87} & \textbf{59.55} & \textbf{58.33}  & \textbf{51.25} & \textbf{0.636}   \\ 
          Mistral-7B &  10.93 & 	18.24  & 14.59 & 25.00 & 48.31 & 54.76  & 42.69 & 0.110   \\  
          Meditron-7B & 8.39  & 	7.22  & 7.81 & \underline{30.43} & 52.81 & 54.76  & \underline{46.00} & 0.072   \\ 
          llm-jp-13B & 15.14  & 	\underline{23.13}  & 19.13 & 28.26 & 37.08 & 51.19 & 38.84 & 0.014 \\ 
          SwallowLM-7B & \underline{19.32}  & 1.15   & 10.24 & 25.00 & 41.57 & 50.00 & 38.86 & 0.056   \\ 
          MMed-Llama3-8B & \textbf{23.00}  & 	17.50  & \textbf{20.25} & 26.09 & \underline{55.06} & \underline{55.95}  & 45.70 & \underline{0.553}   \\ 
        \midrule     
          \multicolumn{9}{|p{5cm}|}{\textbf{Few-shot Evaluation}} \\
        \midrule
          Llama2-7B & 12.89  & 	20.18  & 16.54 & 29.35 & 44.94 & 59.52 & 44.61 & 0.099   \\  
          Llama3-8B &  20.22 & 	28.50  & 24.36 & 34.78 & 53.93 & 63.10  &  50.60 & 0.483   \\ 
          Qwen2-7B & 18.33  &  25.41 & 21.87 & \textbf{44.57} & \underline{56.18} & \textbf{86.90}  & \textbf{62.55} & \textbf{0.625}   \\ 
          Mistral-7B &  12.76 & 	23.05  & 17.91 & 30.43 & \underline{56.18} & 66.67  & 51.09 & 0.378   \\  
          Meditron-7B & 11.79  & 	21.67  & 16.73 & 26.09 & 35.96 & 54.76  & 38.93 & 0.067   \\ 
          llm-jp-13B & \textbf{27.93}  & 	\textbf{28.96}  & \textbf{28.45} & \underline{36.96} & 46.07 & \underline{67.86} & 50.29 &  0.144  \\ 
          SwallowLM-7B &  23.23 &  	23.07 & 23.15 & 30.43 & 44.94 & 59.52  & 44.97 & 0.039 \\ 
          MMed-Llama3-8B & \underline{25.56} & 	\underline{28.73} & \underline{27.14} &  34.78 & \textbf{57.30} & \underline{67.86} & \underline{53.31} & \underline{0.515} \\ 
        \bottomrule
      \end{tabularx}
  \caption{Benchmark results on the rest of other tasks in JMedBench, including Machine Translation (\textbf{EJMMT}), Document Classification (\textbf{CRADE}, \textbf{RRTNM}, \textbf{SMDIS}), and Semantic Text Similarity (\textbf{JCSTS}). The best and second-best performances are highlighted in bold and underlined, respectively.}
  \label{tab:benchmark_rest}
\end{table*}

\subsubsection{Machine Translation}
Table \ref{tab:benchmark_rest} shows the BLEU scores for involved comparison methods on EJMMT. MMed-Llama3-8B and Llama3-8B achieve the best and second-best performance in our benchmark under the zero-shot setting. Interestingly, we find that the English-centric models (e.g., Llama2, Mistral) tend to perform better on translating Japanese texts into English, while the Japanese-centric models (e.g., SwallowLM) perform much better in translating English texts into Japanese. We believe the main reason is the text generation ability in different languages. Therefore, when applying LLMs to the MT task, we should consider more on the language generation ability instead of the language understanding ability. Although the llm-jp is also a Japanese-centric LLM, according to \citet{aizawa2024llm}, it was pre-trained with 50-50 Japanese-English mixed data. Therefore, it has a balanced bilingual NLU and NLG ability. Furthermore, with few-shot demonstrations displaying the task format, llm-jp achieves the best performance in the MT task, which shows the prospect of developing Japanese LLMs from scratch instead of continually pre-training from checkpoints in other languages. Besides, comparing Llama2 and the continually pre-trained Meditron and SwallowLM, we find that continually pre-training with texts in biomedical domains or Japanese texts only will lead to forgetting issues. Continual Learning \cite{wang2024comprehensive} is a potential solution,
but it is still challenging to continually improve the existing LLMs while maintaining their original ability.

\begin{figure*}[htb]
    \centering
    \includegraphics[width=0.9\linewidth]{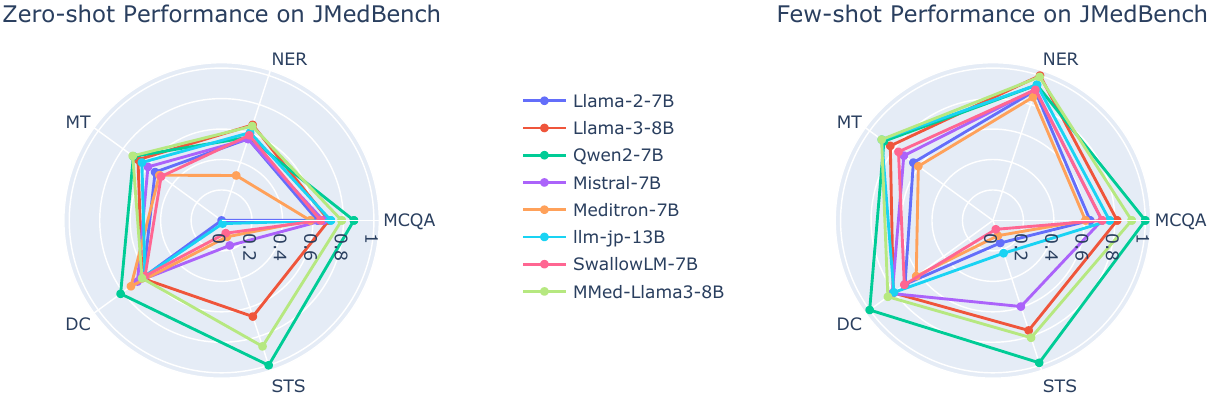}
    \caption{Zero-shot and few-shot performances on different tasks in JMedBench.}
    \label{fig:radar_chart_overall_performance}
\end{figure*}

\subsubsection{Document Classification}
Performances of the DC task are also shown in Table \ref{tab:benchmark_rest}. We find that Qwen2 achieves the best performance again. 
In the zero-shot setting, Meditron achieves the second-best performance, while MMed-Llama3 achieves the second-best performance. Most of the comparison methods achieve better performance when few-shot demonstrations are given. We believe it is because of the provided task format as we discuss in Section \ref{sec:experiment_mcqa}. Moreover, LLMs can also recognize the fine-grained differences between different classes given few-shot demonstrations, making better decisions in classification. Especially, we notice that Meditron performs badly under the few-shot evaluation. We attribute it to the language degradation issue since it accepts a few long documents in the context, amplifying the noise when understanding Japanese.

\subsubsection{Semantic Text Similairity}
The performances on the STS task are varied dramatically. Qwen2 achieves excellent performance on this task, while the prediction of other models like Llama2-based models (i.e., Llama2, Meditron, SwallowLM) is close to random guess. One possible reason is that the distribution of generating numbers is close to a uniform distribution for these models. Recent works also show the shortage of LLMs from this aspect \cite{shah-etal-2023-numeric,avnat2024performance}. However, understanding and generating numbers accurately is essential in the biomedical domains (e.g., on blood test reports). Therefore, it is also a promising search direction in the field of biomedical NLP.

\subsection{Discussions}
In this section, we will conduct an integral and in-depth analysis on the experimental results. 

\subsubsection{Comparison of Model Performances}

Figure \ref{fig:radar_chart_overall_performance} includes two radar charts that demonstrate models' zero-shot and few-shot performance on different tasks. Besides, we also rank the model performance and visualize the rankings in Figure \ref{fig:radar_chart_overall_ranking} as shown in the Appendix. A larger distance from the center represents a higher ranking and better performance. From the radar charts, we can find out that basically, MMed-Llama3, Qwen2, and Llama3 are the most outstanding LLMs on various tasks. Few-shot examples also significantly improve the model performances in all tasks.

\begin{figure}[thb]
    \centering
    \includegraphics[width=\linewidth]{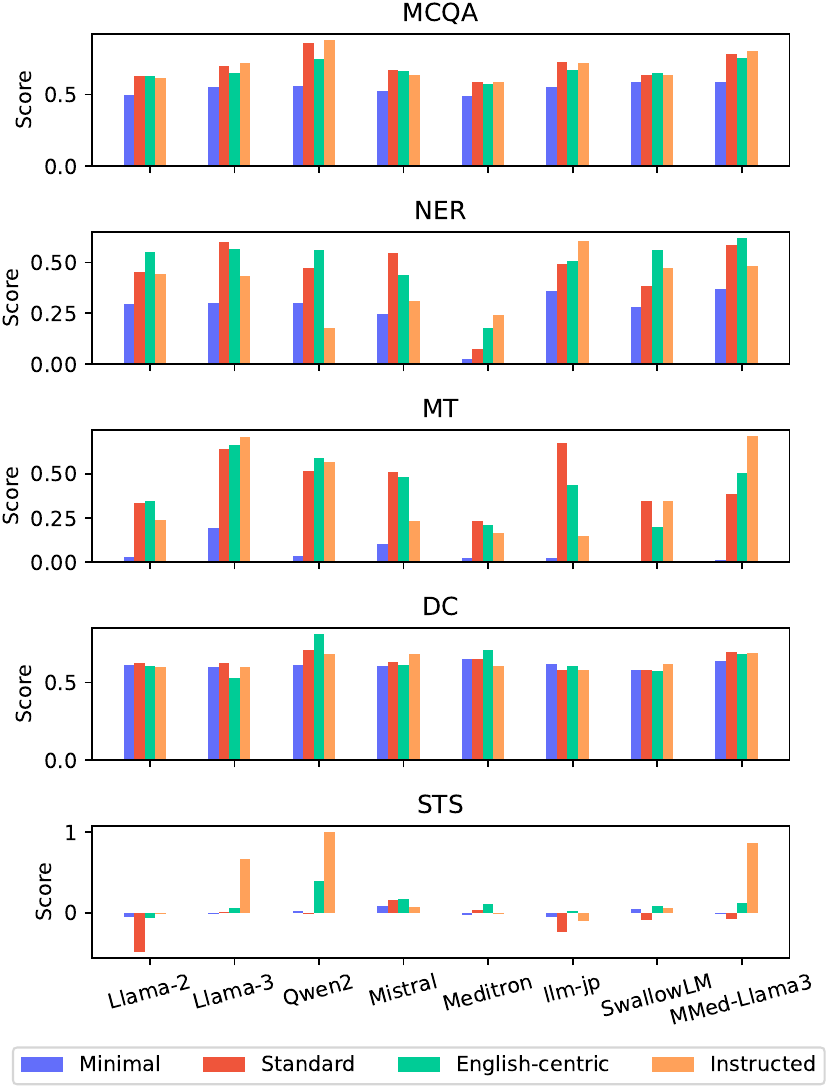}
    \caption{Zero-shot performance under different prompt templates.}
    \label{fig:prompt_template_0}
\end{figure}

\subsubsection{Effect of Prompt Templates}
\begin{figure}[htb]
    \centering
    \includegraphics[width=\linewidth]{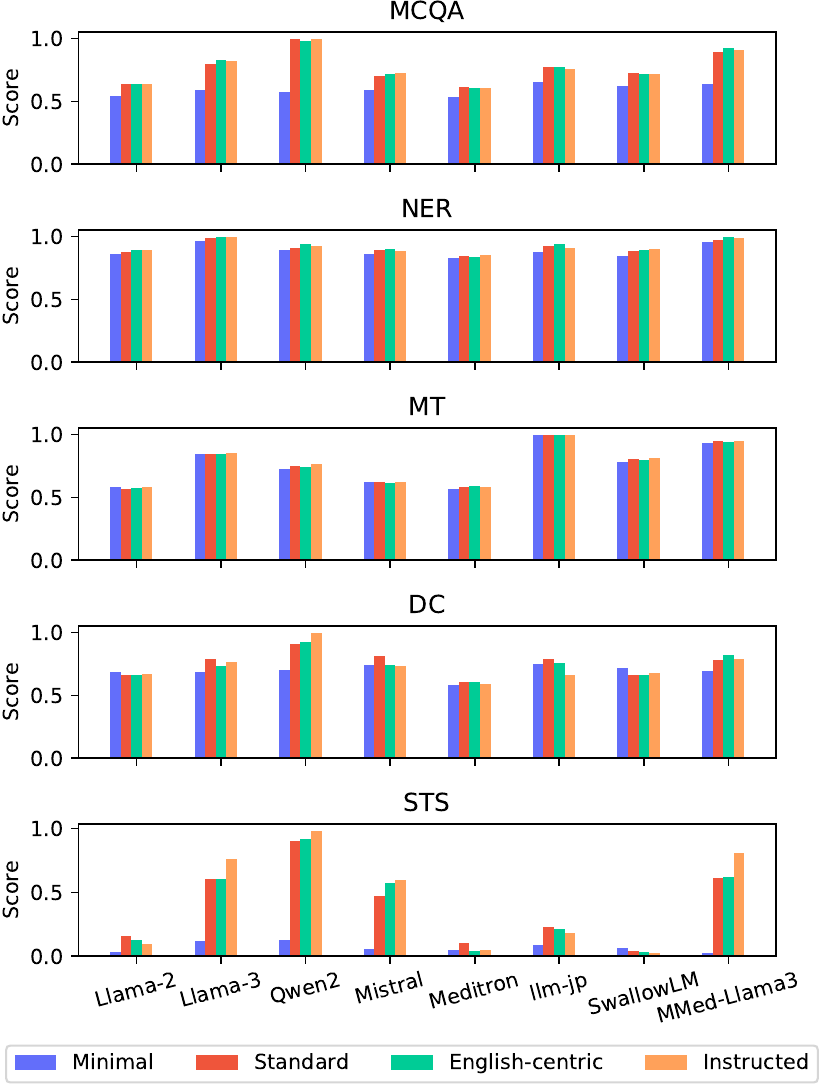}
    \caption{Few-shot performance under different prompt templates.}
    \label{fig:prompt_template_few}
\end{figure}

We also hope to understand the performance of prompt templates across different tasks and models. In zero-shot evaluation, Figure \ref{fig:prompt_template_0} illustrates that the performance of Standard, English-centric, and Instructed prompt templates do not differ significantly, but using English-centric templates usually achieves a slightly better performance. This phenomenon is even more evident in English-centric LLMs. We believe it is because these models have a greater advantage in understanding English instructions, even when facing cross-lingual contexts. Moreover, Figure \ref{fig:prompt_template_few} shows that few-shot demonstrations reduce the differences between prompt templates to a certain extent, with a particularly noticeable enhancement for minimal prompt templates. We believe it is because the output relies less on the instructions and can instead understand the task format from the few-shot examples.

\section{Conclusions}
In this paper, we discuss an urgent need for the field of Japanese biomedical LLMs that requires a solid benchmark for evaluation and comparison. We collect a large collection of Japanese datasets in diverse biomedical tasks, including MCQA, MT, NER, DC, and STS. Considering the scale of the human-manufactured datasets, we translate several large-scale datasets with high quality in English to ensure robust benchmarking results.

Based on the constructed dataset collection, we conduct an evaluation of four types of models, including Japanese biomedical LLMs, Japanese General LLMs, biomedical LLMs in other languages, and general LLMs in other languages. Reported performances reveal some insights for improving existing Japanese LLMs in the biomedical domain. Furthermore, our datasets and evaluation tools are publicly available for future research.


\section*{Limitations}
Considering the difficulty of evaluating natural language generation (NLG) tasks that usually require human evaluation, we only include natural language understanding (NLU) tasks or reformulate NLU tasks into NLU tasks. However, NLG tasks are also widely used in real-world applications. In the future, we consider introducing LLM-based evaluation methods to unlock an easy evaluation of NLG tasks, enriching our benchmark for a further comprehensive evaluation.

Due to the limitation of our budgets, we only translate several datasets of MCQA and NER. We only perform evaluation on models with 7B/8B model parameters. For a comprehensive evaluation, we should also perform comparison in a larger scale. We leave it as a future work to include more translated large-scale datasets in other tasks and evaluation results of larger models. Moreover, though we evaluate these models with four categories of prompt templates, each category only contains one template, which may introduce some fluctuation. To further improve the robustness of our benchmark, we consider including more diverse prompt templates in each prompt category in the future.

Evaluation results on Japanese general domains and biomedical domains in other languages are also valuable for comparison, providing some insights into developing Japanese biomedical LLMs. Such multilingual biomedical benchmark containing diverse tasks is a promising research direction in the future. However, it is out of our scope in this paper.

\section*{Ethics Statement}
We follow the licenses of the involved datasets, which are mainly MIT or CC-BY-4.0\footnote{\url{https://creativecommons.org/licenses/by/4.0/deed.en}}. However, we should note that the NRNER and JCSTS datasets are distributed under the Non-Commercial CC-BY-NC-SA-4.0 license\footnote{\url{https://creativecommons.org/licenses/by-nc-sa/4.0/deed.en}}. In principle, the whole JMedBench should be distributed under a non-commercial license, whereas if it is used for the commercial scenario, these two datasets (i.e., NRNER and JCSTS) should be excluded.

Besides, considering the scale of the existing human-manufactured evaluation datasets, we adopt machine translation system (i.e., GPT-4) to translate some large-scale and high-quality English biomedical datasets into Japanese to fulfill a robust evaluation. However, machine translation system will inevitably generate unfaithful contents. Therefore, those who want to use our datasets to develop faithful biomedical LLMs or biomedical products for real-world application should be aware of this limitation.

\section*{Acknowledgments}

This work was supported by JST SPRING, Grant Number JPMJSP2108 and by Cross-ministerial Strategic Innovation Promotion Program (SIP) on "Integrated Health Care System" Grant Number JPJ012425.

\bibliography{custom}

\appendix
\section{Benchmark Construction Details}
\subsection{Further details of datasets in the JMedBench}
Table \ref{tab:dataset_statistics} shows the statistics of involved datasets in the JMedBench. IgakuQA does not have an official training set, while its genre is similar to MedQA. Therefore, we share the training set of MedQA with IgakuQA for a few-shot evaluation. JMMLU-medical only contains the translated testing set, and we also share the training set of translated MMLU-medical-JP with JMMLU-medical. Considering our limited budgets, we only translated 1,000 training samples randomly selected from the original training set of the PubMedQA. As for the datasets derived from JMED-LLM, including EJMMT, MRNER-Medicine, MRNER-Disease, NRNER, CRADE, RRTNM, and SMDIS, we randomly split a small subset from the original dataset for few-shot evaluation. The size of the training set can be found in Table \ref{tab:dataset_statistics}. As for the JCSTS, we also randomly split a small subset to be the training set. For the rest of the datasets, we strictly follow the origin setting of the split and use the training set or development set for few-show evaluation.

\begin{table}[!htb]
  \tabcolsep=3.5pt
  \centering
  \setlength{\belowcaptionskip}{-.25cm}
  \small
  \renewcommand{\arraystretch}{1.05}
    \resizebox{\linewidth}{!}{
      \begin{tabular}{|>{\centering\arraybackslash}p{1cm}|>{\centering\arraybackslash}p{2.5cm}|>{\centering\arraybackslash}p{1cm}>{\centering\arraybackslash}p{1cm}|>{\centering\arraybackslash}p{1cm}|}
        \toprule
            \textbf{Task} & \textbf{Dataset} & \textbf{Train} & \textbf{Test} & \textbf{Creator} \\
        \midrule
          \multirow{6}{*}{MCQA} & IgakuQA & 10,178 & 989 & Human \\
          & JMMLU-medical  & 45 & 1,271 & Human \\  
          & MedMCQA-JP  & 182,822 & 4,183 & MT \\ 
          & USMLE-QA-JP  & 10,178 & 1,273 & MT \\ 
          & MedQA-JP  & 10,178 & 1,273 & MT \\ 
          & MMLU-medical-JP  & 45 & 1,871 & MT \\ 
          & PubMedQA-JP  & 1,000 & 1,000 & MT \\ 
        \midrule
          MT & EJMMT & 80 & 2,400 & Human \\
        \midrule
          \multirow{7}{*}{NER} & MRNER-Medicine  & 10 & 90 & Human \\  
          & MRNER-Disease  & 10 & 90 & Human \\ 
          & NRNER  & 10 & 90 & Human \\ 
          & BC2GM-JP  & 12,572 & 5,037 & MT \\ 
          & BC5Chem-JP  & 4,562 & 4,801 & MT \\ 
          & BC5Disease-JP  & 4,560 & 4,797 & MT \\ 
          & JNLPBA-JP  & 18,607 & 4,260 & MT \\ 
          & NCBI-Disease-JP  & 5,424 & 940 & MT \\ 
        \midrule
          \multirow{3}{*}{DC} & CRADE  & 8 & 92 & Human \\  
          & RRTNM  & 11 & 89 & Human \\  
          & SMDIS  & 16 & 84 & Human \\  
        \midrule
          STS & JCSTS  & 170 & 3,500 & Human \\ 
        \bottomrule
      \end{tabular}
      }
  \caption{Statistics of involved datasets in JMedBench.}
  \label{tab:dataset_statistics}
\end{table}

\subsection{Prompt Templates for Data Augmentation}
Table \ref{tab:augment_prompt_mcqa} shows the prompt template we used when using OpenAI's APIs for translating biomedical MCQA datasets.

\begin{table*}[hbt]
    \centering
    \small
    \begin{tabular}{p{0.95\linewidth}}
        \toprule
            \textbf{Prompt template for translating MCQA datasets}  \\
        \midrule
            {\tt \#System Message} \\
            {\tt You are an excellent machine translation system for the biomedical domain.} \\
            {\tt Translate Japanese to English.} \\
            {\tt Input and output should be in the same JSON format.} \\
        \midrule
            {\tt \{} \\
            {\tt \quad"question": \{question\}} \\
            {\tt \quad"options": [} \\
            {\tt \quad\quad\{option\_a\},} \\
            {\tt \quad\quad\{option\_b\},} \\
            {\tt \quad\quad\{option\_c\},} \\
            {\tt \quad\quad\{option\_d\},} \\
            {\tt \quad],} \\
            {\tt \quad"context": \{context\} \#Optional} \\
            {\tt \}} \\
        \bottomrule
    \end{tabular}
    \caption{Prompt templates for translating biomedical MCQA tasks.}
    \label{tab:augment_prompt_mcqa}
\end{table*}

Besides, Table \ref{tab:augment_prompt_ner} is the prompt template for translating biomedical NER datasets.

\begin{table*}[htb]
    \centering
    \small
    \begin{tabular}{p{0.95\linewidth}}
        \toprule
            \textbf{Prompt template for translating NER datasets}  \\
        \midrule
            {\tt \#System Message} \\
            {\tt You are an excellent machine translation system for the biomedical domain.} \\
            {\tt Translate Japanese to English.} \\
            {\tt Input and output should be in the same JSON format.} \\
            {\tt Please keep the original key without any changes.} \\
            {\tt Please promise the consistency of translation. For same English words, you should use the same Japanese translation.} \\
            {\tt Please remove unnecessary spaces while translating.} \\
        \midrule
            {\tt \{} \\
            {\tt \quad"entities": \{entities\}}\\
            {\tt \quad"text": \{question\}} \\
            {\tt \}} \\
        \bottomrule
    \end{tabular}
    \caption{Prompt templates for translating biomedical NER tasks.}
    \label{tab:augment_prompt_ner}
\end{table*}

\subsection{Bad Cases during NER Dataset Translation}
\label{app:badcase_ner}
We summarized three main failure types during machine translation: (1) ambiguity of a single word, for example, `depression' can be considered as a mental illness (\textjp{うつ病}) or pressing down (\textjp{抑制}); (2) multiple possible expressions of a single word, for example, `glucose' can be translated into either \textjp{グルコース} or \textjp{血糖}; (3) differences in grammar between English and Japanese. Table \ref{tab:badcase_ner_translation} shows one bad case 
 for each typical failure type during translating NER datasets. The parts underlined indicate an inconsistency between the entity and the text translation. Although there is a small number of failure cases during the machine translation phase, we still realize that the quality of the translation for both the entities and the text is very high during the manual modification process, which can prove the reliability and the scalability of our data augmentation method.

\begin{CJK}{UTF8}{min}
\begin{table*}
    \centering
    \begin{tabular}{m{0.9\linewidth}}
    \toprule
    \textbf{Ambiguity of words}    \\
    \midrule
    \textbf{Original Text}: \underline{Depression} is a major clinical feature of Parkinson' s disease.    \\
    \textbf{Original Entity}: depression \\
    \textbf{Translated Text}: \underline{うつ病}はパーキンソン病の主要な臨床的特徴です。 \\
    \textbf{Translated Entity}: 抑制 \\ 
    \textbf{Explanation}: According to Cambridge English Dictionary, "depression" has multiple meanings: a mental illness (うつ病), or pressing down (抑制). \\
    \midrule
    \textbf{Multiple Expressions of a Single Word}    \\
    \midrule
    \textbf{Original Text}: After recovery from hyperglycaemia, he remained polyuric despite normal blood \underline{glucose} concentrations; water deprivation testing indicated nephrogenic diabetes insipidus, likely to be lithium-induced.     \\
    \textbf{Original Entity}: glucose \\
    \textbf{Translated Text}: 高血糖からの回復後、彼は正常な\underline{血糖}濃度にもかかわらず多尿であり続けました。水制限テストは、リチウム誘発性である可能性のある尿崩症を示しました \\
    \textbf{Translated Entity}: グルコース \\ 
    \textbf{Explanation}: "Glucose" can be translated into either "グルコース" or "血糖". \\
    \midrule
    \textbf{Difference in Grammar}    \\
    \midrule
    \textbf{Original Text}: Molecular cloning and characterization of \underline{two genes encoding gp138}, a cell surface glycoprotein involved in the sexual cell fusion of Dictyostelium discoideum.     \\
    \textbf{Original Entity}: genes encoding gp138 \\
    \textbf{Translated Text}: “Dictyostelium discoideumの性的細胞融合に関与する細胞表面糖タンパク質である\underline{gp138をコードする2つの遺伝子}の分子クローニングと特性評価。 \\
    \textbf{Translated Entity}: gp138をコードする遺伝子 \\ 
    \textbf{Explanation}: Due to grammatical differences, the quantifier "2" is inserted between "genes" and "encoding gp138" when translating the text. \\
    \bottomrule
    \end{tabular}
    \caption{Typical bad cases during NER dataset translation}
    \label{tab:badcase_ner_translation}
\end{table*}
\end{CJK}

\section{Experimental Details}
\label{sec:appendix}

\subsection{Development in chronological order}
\begin{figure*}
    \centering
    \includegraphics[width=\linewidth]{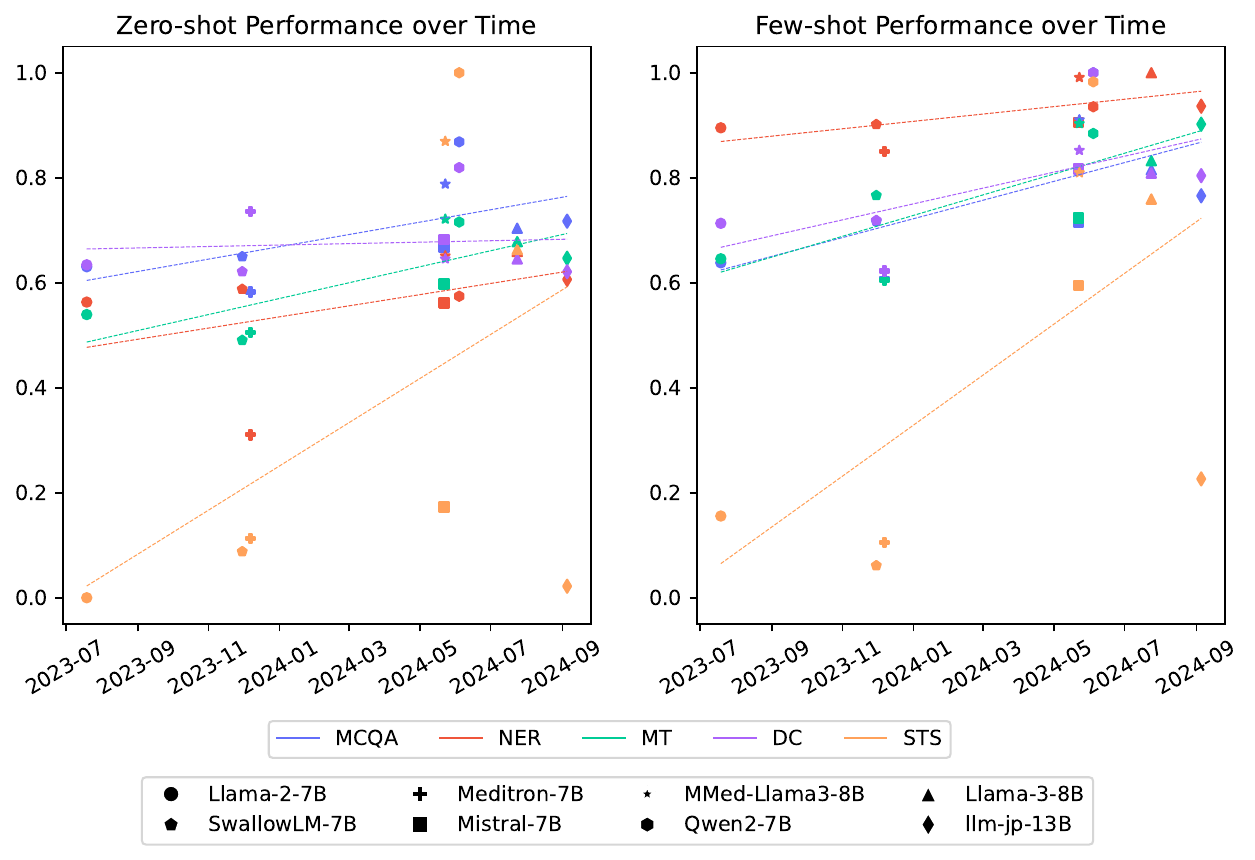}
    \caption{Zero-shot and few-shot performance over time of all involved LLMs.}
    \label{fig:performance_time}
\end{figure*}

We sorted the various models according to their release dates. In chronological order, they are: Llama2-7B (Jul. 2023), SwallowLM-7B (Nov. 2023), Meditron-7B (Dec. 2023), Mistral-7B (May 2024), MMed-Llama3-8B (May 2024), Qwen2-7B (Jun. 2024), Llama3-8B (Jul. 2024), llm-jp-13B (Sep. 2024). Figure \ref{fig:performance_time} illustrates the relationship between model performance and release date. The color of the points represents the corresponding tasks, and the shape represents their models. Colored lines reflect the trend of model performance on each task over time. The figure shows that as time progresses, the performance of models on various tasks is consistently improving, especially for the STS task. Moreover, the improvement in the in-context learning (ICL) capabilities of the models is even more pronounced.

\subsection{Ranking of Models}
\begin{figure*}[htb]
    \centering
    \includegraphics[width=\linewidth]{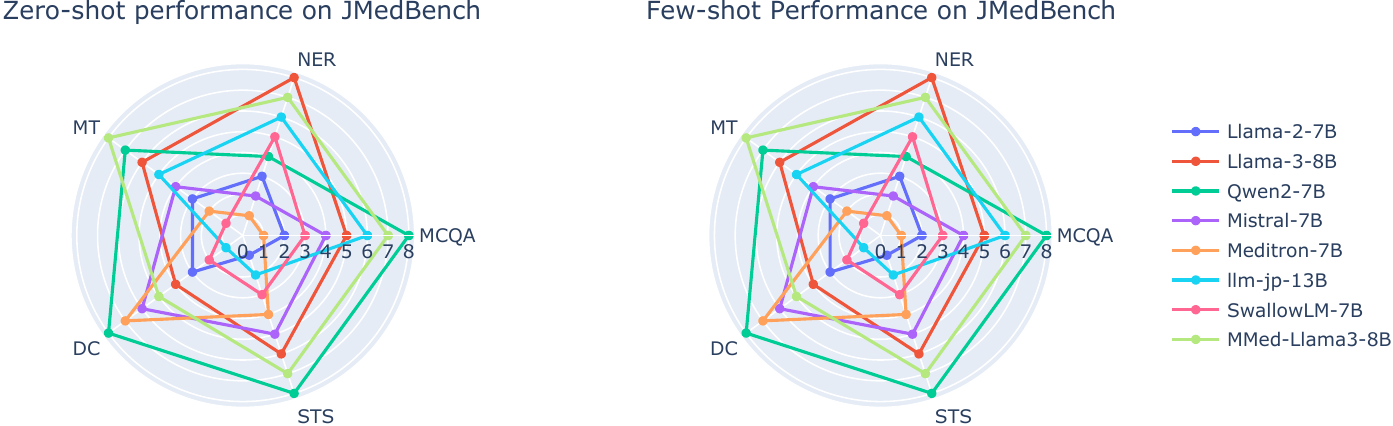}
    \caption{Zero-shot and few-shot performance rankings on JMedBench of all involved LLMs.}
    \label{fig:radar_chart_overall_ranking}
\end{figure*}

Figure \ref{fig:radar_chart_overall_ranking} shows the zero-shot and few-shot performance rankings on JMedBench tasks among all involved LLMs.

\subsection{Comparison Methods}
Detailed information for involved comparison methods is listed in Table \ref{tab:model_info}.

\begin{table*}[bth]
  \tabcolsep=3.5pt
  \centering
  \setlength{\belowcaptionskip}{-.25cm}
  \small
  \renewcommand{\arraystretch}{1.05}
      \begin{tabular}{|>{\centering\arraybackslash}p{5cm}|>{\centering\arraybackslash}p{3cm}|>{\centering\arraybackslash}p{1.3cm}|>{\centering\arraybackslash}p{5cm}|}
        \toprule
            \textbf{Category} & \textbf{Model} & \textbf{\#Params} & \textbf{Checkpoint} \\
        \midrule
          \multirow{5}{*}{General LLMs in other languages} & Llama2-7B & 7B & \texttt{meta-llama/Llama2-7b-hf}  \\
          & Llama3-8B & 8B &  \texttt{meta-llama/Meta-Llama3-8B}   \\
          & Qwen2-7B & 7B &  \texttt{Qwen/Qwen2-7B}   \\
          & Mistral-7B & 7B &   \texttt{mistralai/Mistral-7B-v0.3}  \\
        \midrule
          Biomedical LLMs in other languages & Meditron-7B & 7B &   \texttt{epfl-llm/meditron-7b}  \\
        \midrule
          \multirow{2}{*}{General Japanese LLMs} & llm-jp-13B & 13B &  -   \\
          & SwallowLM-7B & 7B &  \texttt{tokyotech-llm/Swallow-7b-NVE-hf}   \\
        \midrule
          Biomedical Japanese LLMs & MMed-Llama3-8B & 8B &  \texttt{Henrychur/MMed-Llama3-8B}   \\
        \bottomrule
      \end{tabular}
  \caption{Detailed information of involved comparison methods. We contacted the LLM-JP team and used the provided version 3 of the llm-jp model for evaluation.}
  \label{tab:model_info}
\end{table*}

\subsection{Prompts for Each Task}
Detailed prompt templates for each task are shown in Table \ref{tab:prompt_mcqa}, \ref{tab:prompt_ner}, \ref{tab:prompt_mt}, \ref{tab:prompt_dc}, and \ref{tab:prompt_sts}.

\begin{table*}
  \tabcolsep=4pt
  \centering
  \setlength{\belowcaptionskip}{-.25cm}
  \small
      \begin{tabular}{l|p{6cm}|p{6cm}}
        \toprule
            \multicolumn{3}{l}{\textbf{Prompt templates for MCQA task}} \\
        \midrule
             & \textbf{w/o Context} & \textbf{w/ Context}  \\
        \midrule
            \textbf{Minimal} & \texttt{\{question}\} & \texttt{\{context\}} \par \texttt{\{question\}} \par  \\
        
        \midrule
            \textbf{Standard} &  \textjp{質問：}\texttt{\{question\}} \par \texttt{\{options\}} \par \textjp{答え：} &  \textjp{要旨：}\texttt{\{context\}} \par \textjp{質問：}\texttt{\{question\}} \par \textjp{答え：} \\
        \midrule
            \textbf{English-centric} &  \texttt{Question: \{question\}}  \par \texttt{\{options\}} \par \texttt{Answer:} & \texttt{Abstract: \{context\}} \par \texttt{Question: \{question\}}  \par  \texttt{Answer:}\\
        \midrule
            \textbf{Instructed} & \textjp{あなたは医学博士です。基礎科学、臨床科学、医学知識、健康、病気、患者ケア、治療法の基礎となるメカニズムについて理解した上で、以下の選択式問題に答えなさい。以下の選択肢から正しいものを1つ選びなさい。医療ガイドラインに記載されている、現在行われている標準的な治療法に基づいて答えなさい。} \par \textjp{質問：}\texttt{\{question\}} \par \textjp{選択肢：} \par \texttt{\{options\}} \par \textjp{答え：} & \textjp{臨床科学と医学知識の専門家である医師として、次の文が正しいかどうか教えてください。「はい/いいえ/たぶん」のいずれかでお答えください。} \par \textjp{要旨：}\texttt{\{context\}} \par \textjp{質問：}\texttt{\{question\}} \par \textjp{答え：}  \\ 
        \bottomrule
      \end{tabular}
     \caption{Prompt templates for the MCQA task.}
    \label{tab:prompt_mcqa}
\end{table*}

\begin{table*}
  \tabcolsep=4pt
  \centering
  \setlength{\belowcaptionskip}{-.25cm}
  \small
      \begin{tabular}{l|p{12cm}}
        \toprule
            \multicolumn{2}{l}{\textbf{Prompt templates for NER task}} \\
        \midrule
            \textbf{Minimal} & \textjp{段落：}\texttt{\{text\} => \{entity\_type\}}\textjp{：}  \\
        \midrule
            \textbf{Standard} & \textjp{以下の段落において、}\texttt{\{entity\_type\}}\textjp{は？} \par \textjp{段落：}\texttt{\{text\} => \{entity\_type\}}\textjp{：} \\
        \midrule
            \textbf{English-centric} &  Please extract all \texttt{\{entity\_type\}}s mentioned in the paragraph. \par Paragraph: \texttt{\{text\} => \{entity\_type\}:} \\
        \midrule
            \textbf{Instructed} &   \textjp{あなたは医療分野の専門家です。} \par \textjp{あなたは}\texttt{\{entity\_type\}}\textjp{のフレーズを含む段落を与えられます。} \par \textjp{あなたのタスクは段落からこれらすべてのフレーズを抽出することです。} \par \textjp{抽出されたフレーズのみを返し、それらを英語のカンマ（,）で区切る必要があります。} \par
            \textjp{段落：}\texttt{\{text\} => \{entity\_type\}}\textjp{：} \\
        \bottomrule
      \end{tabular}
     \caption{Prompt templates for the NER task.}
    \label{tab:prompt_ner}
\end{table*}

\begin{table*}
  \tabcolsep=4pt
  \centering
  \setlength{\belowcaptionskip}{-.25cm}
  \small
      \begin{tabular}{l|p{6cm}|p{6cm}}
        \toprule
            \multicolumn{3}{l}{\textbf{Prompt templates for MT task}} \\
        \midrule
             & \textbf{English$\rightarrow$Japanese} & \textbf{Japanese$\rightarrow$English}  \\
        \midrule
            \textbf{Minimal} & \texttt{\{source\_text\} => } & \texttt{\{source\_text\} => }  \\
        \midrule
            \textbf{Standard} &  \textjp{翻訳（}\texttt{English => }\textjp{日本語）：}\texttt{\{source\_text\} => } & \texttt{Translation (}\textjp{日本語} \texttt{=> English): \{source\_text\} => } \\
        \midrule
            \textbf{English-centric} & \texttt{Translation (Japanese => English): {source\_text} =>} & \texttt{Translation (English => Japanese): {source\_text} =>}  \\
        \midrule
            \textbf{Instructed} & \textjp{あなたは生物医学文書を翻訳する医学博士です。基礎科学、臨床科学、医学知識、健康、病気、患者ケア、治療法の基礎となるメカニズムを理解した上で、以下の英文を和訳しなさい。} \par \texttt{\{source\_text\} => } & \textjp{あなたは生物医学文書を翻訳する医学博士です。基礎科学、臨床科学、医学知識、健康、病気、患者ケア、治療法の基礎となるメカニズムを理解した上で、以下の和文を英訳しなさい。} \par \texttt{\{source\_text\} => }  \\ 
        \bottomrule
      \end{tabular}
     \caption{Prompt templates for the MT task.}
    \label{tab:prompt_mt}
\end{table*}

\begin{table*}
  \tabcolsep=4pt
  \centering
  \setlength{\belowcaptionskip}{-.25cm}
  \small
      \begin{tabular}{l|p{12cm}}
        \toprule
            \multicolumn{2}{l}{\textbf{Prompt templates for DC task}} \\
        \midrule
            \textbf{Minimal} & \texttt{\{document\}} \par \texttt{\{question\}}  \par \\
        
        \midrule
            \textbf{Standard} &   \textjp{文脈：}\texttt{\{document\}} \par \textjp{質問：}\texttt{\{question\}} \par \texttt{\{classes\}} \par \textjp{答え：} \\
        \midrule
            \textbf{English-centric} &   \texttt{Context: \{document\}} \par \texttt{Question: \{question\}}  \par \texttt{\{classes\}}  \par  \texttt{Answer:}\\
        \midrule
            \textbf{Instructed} &  \textjp{あなたは医学博士です。基礎科学、臨床科学、医学知識、健康、病気、患者ケア、治療法の基礎となるメカニズムについて理解した上で、以下の選択式問題に答えなさい。以下の選択肢から正しいものを1つ選びなさい。} \par \textjp{文脈：}\texttt{\{document\}} \par \textjp{質問：}\texttt{\{question\}} \par \textjp{選択肢：}\texttt{\{classes\}} \par \textjp{答え：}  \\ 
        \bottomrule
      \end{tabular}
     \caption{Prompt templates for the DC task.}
    \label{tab:prompt_dc}
\end{table*}

\begin{table*}
  \tabcolsep=4pt
  \centering
  \setlength{\belowcaptionskip}{-.25cm}
  \small
      \begin{tabular}{l|p{12cm}}
        \toprule
            \multicolumn{2}{l}{\textbf{Prompt templates for STS task}} \\
        \midrule
            \textbf{Minimal} & \texttt{\{text\_1\}} \par \texttt{\{text\_2\}} \par \\
        \midrule
            \textbf{Standard} &   \textjp{テキスト1：}\texttt{\{text\_1\}} \par \textjp{テキスト2：}\texttt{\{text\_2\}} \par \textjp{類似度} \texttt{ (0-5)}\textjp{：} \\
        \midrule
            \textbf{English-centric} &   \texttt{Text 1: \{text\_1\}} \par \texttt{Text 2: \{text\_2\}}  \par \texttt{Semantic Text Similarity (0-5):} \\
        \midrule
            \textbf{Instructed} &  \textjp{あなたは医学博士です。基礎科学、臨床科学、医学知識、健康、病気、患者ケア、治療法の基礎となるメカニズムについて理解した上で、次の2つの文の意味的類似度を0から5の範囲で判断してください。} \par \textjp{0：二つの文は完全に似ていない。} \par \textjp{5：二つの文は完全に同等で、意味が同じである。}  \par \textjp{テキスト1：}\texttt{\{text\_1\}} \par \textjp{テキスト2：}\texttt{\{text\_2\}} \par \textjp{類似度} \texttt{ (0-5)}\textjp{：}  \\ 
        \bottomrule
      \end{tabular}
     \caption{Prompt templates for the STS task.}
    \label{tab:prompt_sts}
\end{table*}

    
    
    
    
    
    
    
    


\end{document}